\documentclass{article}

\usepackage[final]{corl_2019} % Uncomment for the camera-ready ``final'' version
\usepackage{mathtools,fancyhdr,amsmath,amssymb,amsthm,enumitem}

\usepackage{graphicx}
\usepackage{wrapfig}
\usepackage[small]{caption}

\setlist{leftmargin=4.2mm} % Erdem: minimum 4.2mm
\usepackage[bottom]{footmisc}
\usepackage{titlesec}
\titlespacing*{\subsection}{0pt}{0.002\baselineskip}{0.001\baselineskip}
\titlespacing*{\section}{0pt}{0.01\baselineskip}{0.005\baselineskip}
\newcommand{\argmin}{\operatornamewithlimits{arg\,min}}
\newcommand{\argmax}{\operatornamewithlimits{arg\,max}}
\DeclarePairedDelimiter\abs{\lvert}{\rvert}
\DeclarePairedDelimiter\norm{\lVert}{\rVert}

\newcommand{\asymeq}{\stackrel{\boldsymbol{\cdot}}{=}}
\newcommand\mydots{\hbox to 1em{.\hss.\hss.}}

\newtheorem{theorem}{Theorem}

\newtheorem{example}{Example}

\usepackage{xcolor}

\title{Asking Easy Questions: A User-Friendly \\Approach to Active Reward Learning}

\author{
  Erdem B{\i}y{\i}k\\
  Electrical Engineering\\
  Stanford University, USA\\
  \texttt{ebiyik@stanford.edu} \\
  \And
  Malayandi Palan\\
  Computer Science\\
  Stanford University, USA\\
  \texttt{malayandi@stanford.edu} \\
  \And
  Nicholas C. Landolfi\\
  Computer Science\\
  Stanford University, USA\\
  \texttt{lando@stanford.edu} \\
  \And
  Dylan P. Losey\\
  Computer Science\\
  Stanford University, USA\\
  \texttt{dlosey@stanford.edu} \\
  \And
  Dorsa Sadigh\\
  Computer Science and Electrical Engineering\\
  Stanford University, USA\\
  \texttt{dorsa@cs.stanford.edu} \\
}

\begin{document}
\maketitle

%===============================================================================

\vspace{-2em}
\begin{abstract}
Robots can learn the right reward function by querying a human expert. Existing approaches attempt to choose questions where the robot is most uncertain about the human's response; however, they do not consider how \textit{easy} it will be for the human to answer! In this paper we explore an information gain formulation for optimally selecting questions that naturally account for the human's ability to answer. Our approach identifies questions that optimize the trade-off between robot and human uncertainty, and determines when these questions become redundant or costly. Simulations and a user study show our method not only produces easy questions, but also ultimately results in faster reward learning.
\end{abstract}
\vspace{-5px}

% Outline
%%% Introduction
%%% Related Work
%%% Active Learning (Problem Formulation)
%%% Volume Removal (With Example)
%%% Mutual Information
%%%%% Formulation (return to Example where Volume Removal Fails)
%%%%% Works with Same Tricks as Volume Removal (Like Batch)
%%%%% Submodularity (greedy bounded regret proof)
%%%%% Easy (side effect)
%%% Optimal Stopping
%%% Experiments
%%%%% Human Model
%%%%% Simulations
%%%%% User Study

\section{Introduction}
\label{sec:introduction}
Before we deploy a robot, that robot must understand how we want it to behave. Consider a household robot tasked with picking up plates from the kitchen table: when reaching for these plates, how far should the robot stay from obstacles? Would we rather the robot move quickly or slowly? And should the robot take precautions in case it accidentally drops a plate? Often the right behavior---and the reward function that encodes this behavior---is highly personal, and varies from one user to another. A natural way for robots to learn what their current end-user wants is by asking \textit{questions}.

\begin{figure}[t]
	\centering
	\includegraphics[width=0.9\textwidth]{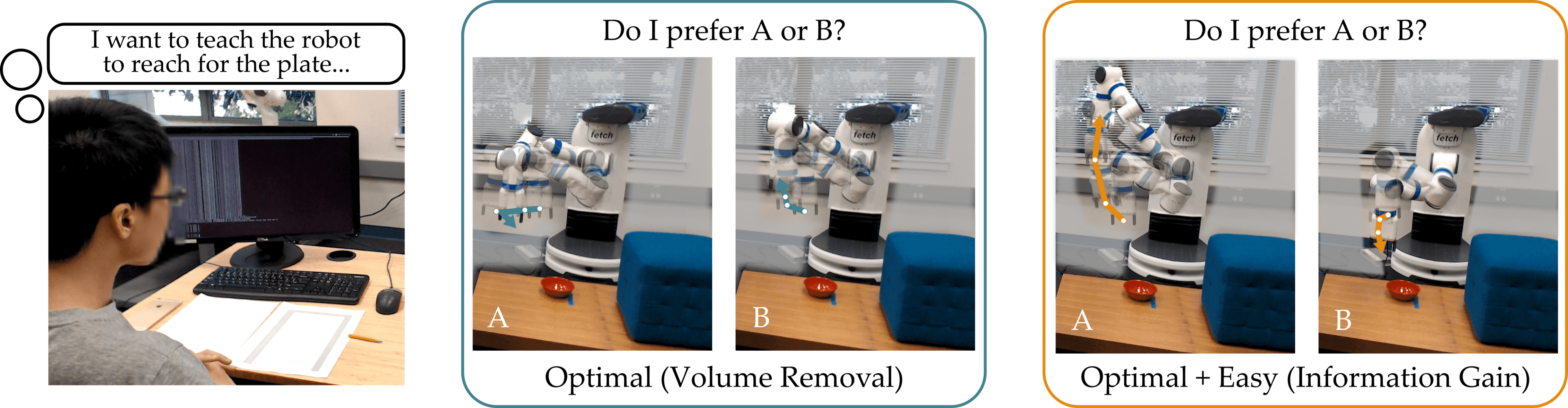}
	\vspace{-5px}
	\caption{Comparing questions that na\"ively minimize model uncertainty to queries that also account for the human's answer uncertainty. Here the robot is attempting to learn the user's reward function, and demonstrates two possible trajectories. The human tries to select the trajectory that better agrees with their own preferences. While the trajectories produced by volume removal are almost indistinguishable, our approach results in questions that are easy to answer: these easy questions also increase the robot's overall learning efficiency.}
	\vspace{-17px}
	\label{fig:frontfig}
\end{figure}

Within today's state-of-the-art, robots solve an optimization problem to choose which questions they ask~\cite{sadigh2017active,  cui2018active, palan2019learning}. But just because a question is optimal does not mean that the human can correctly answer it! Returning to our household robot example, imagine that the robot wants to determine how quickly it should move. There are a range of possible speeds from $0$ to $10$ m/s, and the robot has a uniform prior over this range. When the robot na\"ively attempts to minimize its uncertainty---for example, by optimizing for volume removal---it selects a query that divides the speeds in half; would you rather reach for the plate at a speed of $4.9$ or $5.1$ m/s? While these answer two choices are \textit{optimal}---in the sense that they remove the most incorrect hypotheses---they are also \textit{indistinguishable}, and so the human end-user cannot correctly describe their reward function to the robot (see Fig.~\ref{fig:frontfig}).

In this paper, we focus on asking questions while accounting for the human's ability to answer them. Considering the human while choosing a question is not only important for the human's ease-of-use, but it also improves the robot's learning by reducing wrong answers and indecision. Our insight is:
\vspace{-0.75em}
\begin{center}
	\textit{When robots choose queries to na\"ively minimize their uncertainty, \\the resulting questions may be very difficult for the human to answer.}
\end{center} \vspace{-0.75em}
Based on this observation, we develop an approach that actively learns the human's reward preferences by asking easy questions. We identify a trade-off: robots should choose queries that balance (a) how much information they gain from a correct answer against (b) the human's ability to answer that question confidently. Overall, by reformulating active preference-based reward learning to account for the human's capabilities, we obtain a \textit{set of tools} for user-friendly questions:

\textbf{Information Gain Leads to Easy Questions.} We demonstrate volume removal---a popular state-of-the-art method for active reward learning---does not generate easy queries (Sec.~\ref{sec:volume_removal}). We next explore an alternate objective: we find that when the robot optimizes for information gain, it naturally prioritizes queries that the human can answer confidently, regardless of what the human's individual preferences are (Sec.~\ref{sec:information_gain}). Importantly, optimizing information gain is not more difficult than volume removal within our settings, since both methods here have the same computational complexity.

\textbf{Asking the Right Number of Questions.} Every question the robot asks has an associated cost; for example, the time it takes the human to answer. Although asking additional questions provides more information about the human's reward function, clearly the robot should not ask unlimited questions. Accordingly, we derive an optimal stopping condition so that the robot only asks questions while their expected value outweighs their cost (Sec.~\ref{sec:optimal_stopping}). Our result minimizes the number of queries that the user must respond to, and we include extensions for personalized, query-dependent costs.

\textbf{Simulations and User Studies.} We experimentally compare our approach to the state-of-the-art (Sec.~\ref{sec:experiments}). Across several simulated environments, we show that using the easy questions generated by information gain ultimately leads to faster robot learning than volume removal, and that our information gain approach asks particularly easy questions at the start of its learning. Users reported that our approach asked easier questions in both simulations and experiments on a Fetch robot.

%Our paper is organized as follows. In Section~\ref{sec:problem_formulation} we formalize the problem, and then in Section~\ref{sec:volume_removal} we demonstrate why state-of-the-art solutions lead to difficult questions. We propose an alternative objective that naturally captures question difficulty in Section~\ref{sec:information_gain}, and determine when to stop asking questions in Section~\ref{sec:optimal_stopping}. We experimentally validate our approach with human users in Section~\ref{sec:experiments}.

\section{Related Work}
\label{sec:related_work}
\textbf{Reward Learning.} There are many works that attempt to learn the correct reward function from human feedback. Inverse reinforcement learning (IRL) recovers the reward from expert \textit{demonstrations}, where the human shows the robot how to perform the task \cite{argall2009survey, abbeel2004apprenticeship,ng2000algorithms,abbeel2005exploration,ziebart2008maximum}. But providing demonstrations is often difficult, particularly when the robot has many degrees of freedom \cite{akgun2012keyframe, bajcsy2018learning}. \textit{Preference-based learning} provides a more user-friendly alternative: here the human is shown a few possible trajectories and then asked to select the best one (or rank each of them) \cite{holladay2016active,sadigh2017active,christiano2017deep,biyik2018batch}. We are interested in leveraging preference-based learning while choosing the queries intelligently, so that we improve \textit{data efficiency} and \textit{user experience}.

\textbf{Active Learning.} Active learning seeks to tackle the data inefficiency by selecting the most informative queries, so that the robot learns as much as possible from only a few questions \cite{settles2009active}. Recent research has applied it to preference-based learning using various optimization objectives to select the queries \cite{christiano2017deep,daniel2014active,akrour2012april,sadigh2017active}. Maximizing the volume removed from the hypothesis space (i.e., \textit{volume removal}) is a popular approach \cite{sadigh2017active,golovin2011adaptive,biyik2018batch,biyik2019green,katz2019learning,palan2019learning,basu2019active}. Outside of preference-based learning, active learning papers approximately maximize the information gained from each question (i.e., \textit{information gain}) \cite{golovin2010near,zheng2005efficient,houlsby2011bayesian,mackay1992information,krishnapuram2005semi,herbrich2003fast}. We build on both, with a focus on asking easy questions.

\textbf{User-Friendly Questions.} Related works found that robots which ask difficult questions can frustrate the user \cite{holladay2016active,amershi2014power}, humans want to be able to express uncertainty \cite{guillory2011simultaneous}, and that there is an inherent cost when asking questions (i.e., time, effort). There are some papers that attempt to address these challenges by asking natural questions \cite{racca2019teacher,cakmak2012designing}, incorporating an additional ``About Equal" option to express uncertainty \cite{basu2018learning, krishnan1977incorporating, guo2010real}, or identifying the optimal time to stop asking questions \cite{dimitrakakis2008cost}. Our approach incorporates all three of these challenges to ask the right number of easy questions.

\section{Problem Formulation}
\label{sec:problem_formulation}
\newcommand{\states}{\mathcal{S}}
\newcommand{\actions}{\mathcal{A}}

\textbf{Model.} We consider a deterministic, fully-observable dynamical system. We use $s_t\in \states$ to denote the state of the system and $a_t\in\actions$ to denote the action (control
input) to the system, both at time $t$. A trajectory, $\xi\in\Xi$, is a finite sequence of states and actions. i.e., $\xi = ((s_t, a_t))_{t=0}^T$, where $T$ is the time horizon of the system. Since the system is deterministic, a trajectory $\xi$ can be more succinctly represented by $\Lambda = (s_0, a_0, a_1, \ldots, a_T)$, the initial state and sequence of actions in the trajectory. We use $\xi$ and $\Lambda$ interchangeably when referring to a trajectory, depending on the context.

We assume a reward function, $R: \Xi \to \mathbb{R}$, describes how a human wants the system to behave. Our goal is to learn $R$. More formally, let $R$ be a linear combination of selected features $\Phi: \Xi \to \mathbb{R}^d$, so that $R(\xi) = \omega^\top \Phi(\xi)$. To learn $R$, we simply have to learn the human's $\omega$.

\textbf{Preference Queries.} A \emph{preference query} $Q:=\{\xi_1,\xi_2,\mydots,\xi_K\}$ is a set of $K$ trajectories; the human responds to a preference query by picking their most preferred trajectory from this set. We will learn the reward parameter  $\omega$ by choosing---and having the human respond to---a sequence of queries.
%(See \cite{sadigh2017active, biyik2018batch, palan2019learning} for more details on the benefits of using preference queries over other forms of human feedback.)

We want to find the true reward parameter using as few preference queries as tractably possible. In other words, we want to find the sequence of queries that maximizes the amount of information that we expect to receive about $\omega$. Such a sequence of queries should recognize there is a human-in-the-loop; it should also generate queries that are ``easy" for the human, thereby minimizing the amount of bad feedback it may receive. Unfortunately, this problem is, in general, NP-hard \cite{ailon2012active}.

We therefore proceed in a greedy fashion: we first initialize a distribution over $\omega$, and then iteratively alternate between (a) generating a single query and (b) using the human's response to that query to update our distribution over $\omega$. For some query generation strategies, such as the maximum volume removal method (see Sec.~\ref{sec:volume_removal}), this greedy approach has bounded regret \cite{sadigh2017active}.

We first focus on the second step: updating our distribution over $\omega$ based on the human's response. Let $P(q \mid Q, \omega)$ denote the probability that the human chooses trajectory $q$ from the query $Q$ when their reward parameter is $\omega$ (see Section~\ref{sec:human_model} for common human choice models). Then, given a prior $p(\omega)$ and the human's response $q$ to the query $Q$, we can compute a posterior over $\omega$:
\begin{align*}
P(\omega \mid Q,q) \propto P(q \mid Q,\omega)P(\omega).
\end{align*}
In the next sections, we discuss methods for generating the queries $Q$ that the human will answer.

\section{Volume Removal Solution}
\label{sec:volume_removal}
Maximizing volume removal is one popular strategy for selecting queries. The method attempts to generate the most-informative queries by finding the query that maximizes the expected difference between the prior and \textit{unnormalized} posterior \cite{sadigh2017active, biyik2018batch, palan2019learning, biyik2019green}. Formally, the method generates a query of $K$ trajectories (for $K \geq 2$) at iteration $n$ by solving:
\vspace{-2px}
\begin{align}
\label{eq:og_old_optimization}
Q^*_n &= \argmax_{Q_n= \{\Lambda_1,\dots,\Lambda_K\}} \mathbb{E}_{q_n} \mathbb{E}_\omega \left[1 - P(q_n\mid Q_n,\omega)\right].
\end{align}\\[-9px]
The distribution over $\omega$ can get very complex and thus---to tractably compute the expectation---we are forced to resort to sampling. Letting $\Omega$ denote a set of $M$ samples from the prior $P(\omega)$ and $\asymeq$ denote asymptotic equality as $M\to\infty$, the optimization problem can be re-written as:
\vspace{-5px}
\begin{align}
Q^*_n \asymeq \argmin_{Q_n=\{\Lambda_1,\dots,\Lambda_K\}}\sum_{q_n\in Q_n}\left(\sum_{{\omega}\in\Omega}P(q_n\mid Q_n,{\omega})\right)^2.
\label{eq:final_old_optimization}
\end{align}

\vspace{-10px}
\textbf{Failure Case.}
Although prior works have shown that volume removal works well in practice, here we reveal that the optimization problem used to identify volume removal queries \textit{does not} capture our original goal of generating informative queries. Specifically, the trivial query, which outputs $K$ \emph{identical} trajectories as options, is in fact a global solution to the volume removal formulation.

\begin{theorem}
	The trivial query, $Q\!=\!\{\xi_A, \xi_A, \mydots, \xi_A\}$ (for any $\xi_A\in\Xi$) is the global solution to \eqref{eq:og_old_optimization}.
	\vspace{-10px}
	\begin{proof}
	For a given $Q$ and $\omega$, $\sum_q P(q\!\mid\!Q,\omega)\!=\!1$. Thus, we can bound the objective in $\eqref{eq:og_old_optimization}$ as follows:
	\vspace{-11px}
	\begin{align*}
		\mathbb{E}_{q_n} \mathbb{E}_\omega [1 - P(q_n\mid Q_n,\omega)] = 1 - \mathbb{E}_w \sum_{q_n\in Q_n}P(q_n\mid Q_n,\omega)^2 \leq 1-1/K.
	\end{align*}\\[-9px]
	For the trivial query $Q\!=\!\{\xi_A, \xi_A, \mydots, \xi_A\}$, the objective in \eqref{eq:og_old_optimization} has value $\mathbb{E}_{q_n} \mathbb{E}_\omega \left[1\!-\!P(q_n\mid Q,\omega)\right]\!=\!1\!-\!1/K$, equal to the upper bound on the objective and thus, it is a global solution to \eqref{eq:og_old_optimization}.
	\end{proof}
	\vspace{-10px}
\label{thm:volume_removal_failure}
\end{theorem}

The robot should ask questions to learn the human's preferences: but when all answer options are the same, the robot gains \emph{no information at all} about what the human prefers! Hence, the maximum volume removal method works well in practice not despite the non-convexity of the optimization problem, but because of it. This non-convexity allows solution methods to converge to other locally optimal queries, which may be more informative than the globally optimal trivial query.

\label{sec:hard_queries}

\textbf{Hard Queries.} Even when volume removal identifies queries that are not identical, the questions it asks may still be very challenging for the user to answer \cite{palan2019learning}. Here we provide a concrete example:

\begin{example}\label{ex:example_question}
	\normalfont Let the robot query the human while providing two answer options, $q_1$ and $q_2$. First, consider a query $Q_A$ such that $P(q_1\! \mid\! Q_A, \omega)\! =\! P(q_2\! \mid\! Q_A, \omega) \ \forall \omega\!\in\!\Omega$. Intuitively, responding to $Q_A$ is  ``hard" for the human, since their probability of selecting either option is equal. Alternatively, consider a query $Q_B$ such that $P(q_1\! \mid\! Q_B, \omega)\! \approx\!  1 \ \forall \omega\! \in\!\Omega^{(1)}$ and $P(q_2\! \mid\! Q_B, \omega)\! \approx\! 1 \ \forall \omega\! \in\! \Omega^{(2)}$, where $\Omega^{(1)}\! \cup\! \Omega^{(2)}\! =\! \Omega$ and $\abs{\Omega^{(1)}}\!=\!\abs{\Omega^{(2)}}$. If the true $\omega$ lies in $\Omega^{(1)}$, the human will always select $q_1$ and conversely, if the true $\omega$ lies in $\Omega^{(2)}$, the human will always select $q_2$. 
Intuitively, this query is ``easy" for the human: regardless of their true reward, the choice is almost certain.
%Intuitively, this is a query which is "easy" for the human to respond to since, regardless of the human's true reward, her choice is almost certain.

An intelligent robot should select $Q_B$ over $Q_A$: not only does $Q_A$ fail to provide information about the human's reward (because their answer could be explained by any $\omega$), but it is also hard for the human to answer (since both options seem equal). When maximizing volume removal, however, the robot thinks $Q_A$ is just as good as $Q_B$: they are both global solutions to its optimization problem. Fig.~\ref{fig:experiment_visuals} demonstrates hard queries generated by the volume removal formulation.
\end{example}

\section{Information Gain Solution}
\label{sec:information_gain}
In this section we outline an alternate objective that robots should use to select useful questions. We focus on addressing the short-comings of volume removal by introducing a trade-off: instead of na\"ively optimizing to minimize the robot's uncertainty, the robot should also consider the human's ability to answer. We find we can utilize \textit{information gain} to incorporate this trade off, encouraging the robot to ask easy but informative queries. Moreover, we demonstrate that leveraging information gain to select queries has the same benefits as volume removal without its potential drawbacks.

Let us first introduce our method. At each step, we attempt to find the query that maximizes the expected information gain about $\omega$. We can do so by solving the following optimization problem: 
\begin{align}\label{eq:og_new_optimization}
Q^*_n &= \argmax_{Q_n=\{\Lambda_1,\dots,\Lambda_K\}} I(\omega; q_n \mid Q_n) = \argmax_{Q_n=\{\Lambda_1,\dots,\Lambda_K\}} H(\omega\mid Q_n) - \mathbb{E}_{q_n} H(\omega \mid q_n,Q_n),
\end{align}\\[-7px]
where $I$ is the mutual information and $H$ is Shannon's information entropy \cite{cover2012elements}. By approximating the expectations via sampling, we re-write this optimization problem as:
\vspace{-3px}
\begin{align}\label{eq:final_new_optimization}
Q^*_n &\asymeq \argmax_{Q_n=\{\Lambda_1,\dots,\Lambda_K\}} \frac{1}{M} \sum_{q_n\in Q_n}\sum_{{\omega}\in\Omega}P(q_n \mid Q_n,{\omega})\log_2{\left(\frac{M \cdot P(q_n\mid Q_n,{\omega})}{\sum_{\omega'\in\Omega}P(q_n \mid Q_n,\omega')}\right)}.
\end{align}\\[-7px]
Robots that solve \eqref{eq:final_new_optimization} to select their questions do not show identical trajectories to the user: the trivial query $Q = \{\xi_A,\ldots, \xi_A\}$ is not the global solution to our new optimization problem. Instead, such a query is the actually \emph{global minimum}. Additionally, the complexity of computing objective \eqref{eq:final_new_optimization} is equivalent (in order) to the maximum volume removal objective \eqref{eq:final_old_optimization}.
Thus, while being \emph{at least as tractable}, our method avoids the failure case of the volume removal formulation (Theorem~\ref{thm:volume_removal_failure}).

\subsection{Generating Easy Queries}
\label{sec:easiness}
Unlike volume removal, our information gain formulation reasons over the trade-off between two sources of uncertainty: robot and human. To see this, we re-write the optimization problem as:
\begin{align}
Q^*_n &= \argmax_{Q_n=\{\Lambda_1,\dots,\Lambda_K\}} H(q_n \mid Q_n) - \mathbb{E}_{\omega} H(q_n \mid \omega, Q_n).
\label{eq:easiness_equation}
\end{align}
\vspace{-10px}

\begin{figure}[t]
	\centering
	\includegraphics[width=0.7\textwidth]{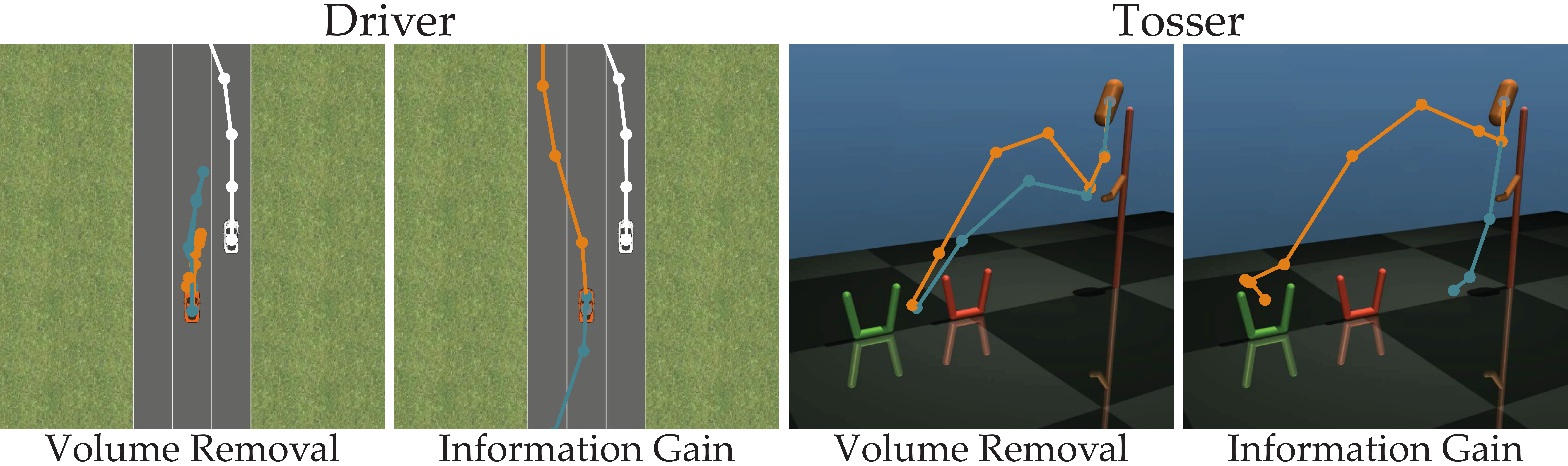}
	\vspace{-4px}
	\caption{Sample queries generated with the volume removal and information gain methods on 2 different tasks. Volume removal generates queries that are difficult, because the options are almost equally good.}
	\vspace{-15px}
	\label{fig:experiment_visuals}
\end{figure}
Here, the first term is the \textit{robot's uncertainty} over the human's response: i.e., how well can the robot predict what the human will answer. The second term is the \textit{human's uncertainty} when answering: i.e., given that the user has reward parameter $\omega$, how confidently does she choose option $q$? Optimizing for information gain, without any additional easiness term, considers both types of uncertainty, and favors questions where (a) the robot is unsure how the human will answer but (b) the human can answer easily. We visualize some example queries produced by information gain in Figs.~\ref{fig:frontfig} and \ref{fig:experiment_visuals}. We also concretely show our objective favors easy queries by returning to Example~\ref{ex:example_question} from Sec.~\ref{sec:hard_queries}.

\begin{example}\normalfont The first term in Eq.~\eqref{eq:easiness_equation} has the same value for both $Q_A$ and $Q_B$. However, $Q_A$ attains the global maximum of the second term while $Q_B$ attains the global minimum! Thus, the overall value of $Q_B$ is higher and, as desired, the robot recognizes that $Q_B$ is a more intelligent question.
\end{example}

\subsection{Theoretical Guarantees}
\label{sec:theoretical_guarantees}

When comparing our method to volume removal, we avoid its two pitfalls of identical options (Theorem~\ref{thm:volume_removal_failure}) and difficult questions (Example \ref{ex:example_question}). But we also inherit many of the same benefits, including an assurance that greedy optimization leads to bounded regret. Recall that---for both volume removal and information gain---the robot selects its question without considering the future questions. \citet{sadigh2017active} have demonstrated, by virtue of the maximum volume removal objective being submodular, this greedy algorithm is guaranteed to have bounded sub-optimality in terms of the volume removed. Our method is greedy as well, and below we show it enjoys similar theoretical guarantees.

\begin{theorem}
	We assume that the queries are generated from a finite set and we ignore any errors due to sampling. Then, the sequence of queries generated according to \eqref{eq:final_new_optimization} at each iteration is at least $1-\epsilon$ times as informative as the optimal sequence of queries after $\log \frac{1}{\epsilon}$ as many iterations.
	\vspace{-12px}
	\begin{proof}
		In Eq.~\eqref{eq:final_new_optimization}, we aim to maximize the mutual information between $\omega$ and the set of queries $S$, which is monotone in $S$. Recall that the mutual information is a submodular function. Therefore, our objective is a submodular, monotone function, and the desired result follows directly from \cite{nemhauser1978analysis}.
	\end{proof}
	\vspace{-12px}
\end{theorem}

\subsection{Extensions}
\vspace{-2px}
Building off the volume removal method, many extensions have been developed such as batch optimization \cite{biyik2018batch,biyik2019batch} or warm starting techniques \cite{palan2019learning}. These extensions are agnostic to the details of the optimization; they simply require the query generation algorithm operates in a greedy manner and maintains a distribution over $\omega$. Hence, these extensions can be readily applied to our method.

Thus, when compared to the maximum volume removal, our method (1) has the same theoretical guarantees; (2) has the same computational complexity; (3) does not have a fundamental failure case; (4) generates easier queries for the human to respond; and (5) can leverage the same extensions.

\section{Optimal Stopping}
\label{sec:optimal_stopping}
\vspace{-4px}
In the previous section, we showed how to improve the preference-based learning by easy and informative questions. We now further improve user-experience with an automatic stopping criterion: the active learning should end when the robot's questions get more costly than they are informative.

We first extend the preference-based learning framework so that each query $Q$ has an associated cost $c(Q)$. This function captures the \textit{cost} of a question: the amount of time it takes for the human to answer, the number of similar questions that the human has already seen, or even the interpretability of the question itself. We next subtract this cost from our information gain objective (\ref{eq:og_new_optimization}), so that the robot maximizes information gain while biasing its search towards low-cost questions:
\vspace{-3px}
\begin{align}
r^*_n = \max_{Q_n=\{\Lambda_1,\dots,\Lambda_K\}} I(\omega; q_n \mid Q_n) - c(Q_n).
\label{eq:stopping_opt}
\end{align}\\[-8px]
Now that we have introduced a cost into the query selection problem, the robot can reason about when its questions are becoming prohibitively expensive or redundant. Intuitively, the robot should stop asking questions if their cost outweighs their value; and, indeed, that is the optimal stopping condition when the robot maximizes information gain.

\begin{theorem}\label{thm:optimal_stopping}
	A robot using information gain to perform active preference-based learning should stop asking questions if and only if the global solution to \eqref{eq:stopping_opt} is negative at the current timestep.
	\vspace{-7px}
	% \begin{proof}
	% 	We need to show if the global optimum is negative, then any longer-horizon optimization will also give negative reward in expectation. Let $Q^*_n$ denote the global optimizer. For any $k\geq0$,
	% 	\begin{align*}
	% 	I(&q_n,\dots,q_{n+k} ; \omega | Q_n,\dots,Q_{n+k},\mathcal{D}^{(n-1)}) - \sum_{i=0}^{k}c(Q_{n+i}) \\
	% 	&= I(q_n ; \omega | Q_n,\mathcal{D}^{(n-1)})\! +\! \mydots\! +\! I(q_{n+k} ; \omega | q_n, \mydots, q_{n+k-1}, Q_n,\mydots,Q_{n+k},\mathcal{D}^{(n-1)})\! -\! \sum_{i=0}^{k}c(Q_{n+i})\\
	% 	&\leq I(q_n ; \omega | Q_n,\mathcal{D}^{(n-1)})\! +\! \mydots\! +\! I(q_{n+k} ; \omega | Q_{n+k},\mathcal{D}^{(n-1)})\! -\! \sum_{i=0}^{k}c(Q_{n+i})\\
	% 	&\leq (k+1)\left[I(q_n ; \omega | Q^*_n,\mathcal{D}^{(n-1)}) - c(Q^*_n)\right] < 0
	% 	\end{align*}
	% 	where the first inequality is due to the submodularity of the mutual information, and the second inequality is because $Q^*_n$ is the global maximizer of the greedy objective. The other direction of the proof is very clear: If the global optimizer is nonnegative, then querying $Q^*_n$ will not decrease the cumulative active learning reward in expectation, so stopping is not optimal.
	% \end{proof}
\end{theorem}The proof of the result is presented in the Appendix. 

The decision to terminate our learning algorithm is straightforward: at each timestep, if $r^*_n$ is non-negative, we ask the human to respond to another query; otherwise, the robot stops asking questions. This automatic stopping procedure helps make the active learning process more user-friendly by ensuring that the user does not have to respond to any unnecessary or redundant queries.

%\begin{corollary}
%	Theorem~\ref{thm:optimal_stopping} is still valid if weak preference queries are used.
%	\label{cor:weak_preferences}
%\end{corollary}
% \begin{corollary}
% 	For an extended version of volume removal formulation with query-dependent costs by simply subtracting $c(Q_n)$ in the objective of Eq.~\eqref{eq:final_old_optimization}, Theorem~\ref{thm:optimal_stopping} holds, because the volume removal is also submodular.
	%Similar to Corollary~\ref{cor:weak_preferences}, the theorem is also valid with weak preferences.
% \end{corollary}

\section{Experiments}
\label{sec:experiments}
We argue that our information gain approach for asking easy questions is not only user-friendly, but also improves the robot's learning rate. In this section we experimentally compare volume removal and information gain: we begin in Sec.~\ref{sec:human_model} by defining how the robot models the user, and then we review our results across simulated environments (Sec.~\ref{sec:simulations}) and a user study (Sec.~\ref{sec:user_study}).\footnote{The code is available at \url{http://github.com/Stanford-ILIAD/easy-active-learning}.}
%We experimented with a variety of dynamical systems in simulation as well as on a Fetch mobile manipulator platform.

\subsection{Human Choice Model}
\textbf{Standard Model.}
\label{sec:human_model}
We require a probabilistic model for the human's choice $q$ in a query $Q$ conditioned on their reward parameters $\omega$. Previous work demonstrated the importance of modeling imperfect human responses  \cite{kulesza2014structured}. We model a noisily rational human as selecting $\xi_k$ from $Q$ by
\begin{align}
P(q=\xi_k\mid Q,\omega) &= \frac{\exp(R(\xi_k))}{\sum_{\xi\in Q}\exp(R(\xi))}.%, \forall \xi_k\in Q.
\label{eq:noisy_rational}
\end{align}
This model, backed by neuroscience and psychology \cite{daw2006cortical,luce2012individual,ben1985discrete,lucas2009rational},  is routinely used  \cite{biyik2019green, guo2010real, viappiani2010optimal}.

\textbf{Extended Model.}
We generalize this preference model to include an ``About Equal" option for queries between two trajectories.
We denote this option by $\Upsilon$ and define a \emph{weak preference query} (as opposed to strict preference queries) $Q^+ := Q\cup\{\Upsilon\}$ when $K=2$.

%Prior work has proposed a hard threshold based on trajectory feature values \cite{basu2018learning,guo2010real}. comment from nick: probably don't need this here, seems like related work
%Moreover, although such a response would give the information that $R(\xi_1)$ and $R(\xi_2)$ are close to each other, this kind of information is often neglected by ignoring that question \cite{christiano2017deep,basu2018learning}. 
Building on prior work \cite{krishnan1977incorporating}, we incorporate the information from the ``About Equal" option by introducing a minimum perceivable difference parameter $\delta \geq 0$, and defining:
\begin{align}
P(q=\Upsilon \mid Q^+,\omega) &= \left(\exp(2\delta) - 1\right)P(q=\xi_1\mid  Q^+,\omega)P(q=\xi_2 \mid  Q^+,\omega)\:, \nonumber\\
P(q=\xi_k\mid Q^+,\omega) &= \frac{1}{1+\exp(\delta + R(\xi_{k'}) - R(\xi_k))}, \forall \xi_k,\xi_{k'}\in Q^+\setminus\{\Upsilon\}, k\neq k'.
\label{eq:weak_noisy_rational}
\end{align}
Notice that Eq.~\eqref{eq:weak_noisy_rational} reduces to Eq.~\eqref{eq:noisy_rational} when $\delta=0$; in which case we model the human as always perceiving the difference in options.
All derivations in earlier sections hold with weak preference queries.
In particular, we include a discussion of extending our formulation to the case where $\delta$ is user-specific and unknown in the Appendix. 
The additional parameter causes no trouble in practice. 

We note that there are alternative choice models compatible with our framework for weak preferences (e.g., \cite{holladay2016active}).
Additionally, one may generalize the weak preference queries to $K>2$, though it complicates the choice model as the user must specify which of the trajectories create uncertainty.

\subsection{Simulations}
\label{sec:simulations}

We perform experiments in a variety of environments: (1) a linear dynamical system (\textbf{LDS)} with six-dimensional state and three-dimensional action space; (2) A two-dimensional \textbf{Driver} environment \cite{sadigh2016planning} which simulates an autonomous car with steering, braking and acceleration capabilities driving in an environment with another vehicle;
% and one other car with a fixed trajectory; 
 (3) A \textbf{Tosser} robot built in MuJoCo \cite{todorov2012mujoco} that tosses capsule-shaped object into baskets; (4) A \textbf{Fetch} mobile manipulator robot \cite{wise2016fetch} (using OpenAI Gym \cite{brockman2016openai} for simulation) that performs a reaching task among obstacles.
Fig.~\ref{fig:experiment_visuals} visualizes Driver and Tosser.

%We perform experiments in different simulation environments:
%\begin{itemize}[nosep]
%	\item A linear dynamical system (LDS) with 6D state space and 3D action space,
%	\item A 2D driving simulator \cite{sadigh2016planning}, that simulates an autonomous car with $2$ inputs (steering and acceleration) and another car with a fixed trajectory,
%	\item A Tosser robot built on MuJoCo \cite{todorov2012mujoco}, that tosses a capsule-shaped object into baskets, and
%	\item A Fetch mobile manipulator robot simulation \cite{wise2016fetch} from OpenAI Gym \cite{brockman2016openai} again built on MuJoCo, that simulates a robot trying to reach an object while avoiding an obstacle.
%\end{itemize}

\textbf{Implementation.} We adopt the features and action spaces from \cite{biyik2019batch} (detailed in Appendix).
We normalize features to have unit variance under uniformly random actions. We set $K\!=\!2$, and when using the ``About Equal" option set $\delta\!=\!1$.
To accelerate query generation, we generate $500,000$ queries with uniformly random actions and optimize over this finite set.
To judge convergence of inferred reward parameters to true parameters, we adopt the \emph{alignment metric} from \cite{sadigh2017active}, $m\!=\!\frac{1}{M}\sum_{\bar{\omega}\in\Omega}\frac{\omega^\top\bar{\omega}}{\norm{\omega}_2\norm{\bar{\omega}}_2}$
with $M\!=\!100$, and $\Omega$ sampled using Metropolis-Hastings algorithm.

% if we want to keep this structure leaving here in case.
%\noindent\textbf{H1.} Information gain formulation outperforms the volume removal in terms of data-efficiency.\\
%\noindent\textbf{H2.} The queries of the information gain formulation are easier than those of the volume removal.\\
%\noindent\textbf{H3.} Using the information from ``About Equal" responses increases the performance.\\
%\noindent\textbf{H3.} Optimal stopping enables cost-efficient reward learning under various costs.

\noindent\textbf{Hypothesis 1.} \emph{Information gain formulation outperforms volume removal w.r.t. data-efficiency}.

\begin{figure}[t]
	\centering
	\includegraphics[width=\textwidth]{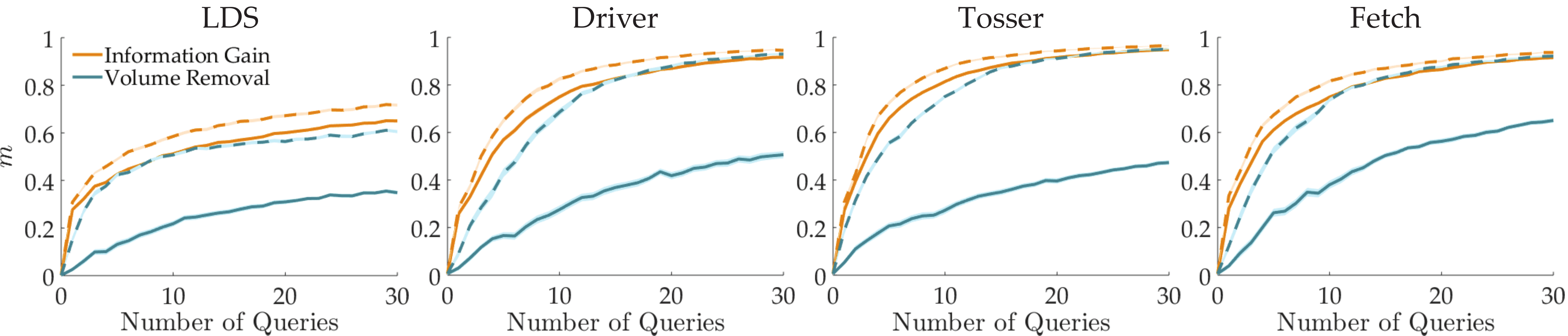}
	\vspace{-15px}
	\caption{Alignment values are plotted (mean$\pm$s.e.). Dashed lines show the weak preference query variants.}
	\vspace{-10px}
	\label{fig:simulation_results_m}
\end{figure}

We sample $100$ reward parameters uniformly at random from $\{\omega\!\in\!\mathbb{R}^d \!\mid\! \norm{\omega}_2\!=\!1\}$. We learn the reward functions via both strict and weak preference queries, using simulated responses. Fig.~\ref{fig:simulation_results_m} shows the alignment value against query number for the $4$ different tasks. Even though the ``About Equal" option improves the performance of volume removal by preventing $Q=\{\xi_A,\xi_A,\dots\}$ from being a global optimum, information gain gives a significant improvement on the learning rate both with and without the ``About Equal" option in all environments.\footnote{See the Appendix for  results without query space discretization.} These results strongly support \textbf{H1}.

\noindent\textbf{Hypothesis 2.} \emph{Information gain queries are easier than those from volume removal}.
\vspace{-4px}

\begin{figure}[t]
	\centering
	\includegraphics[width=\textwidth]{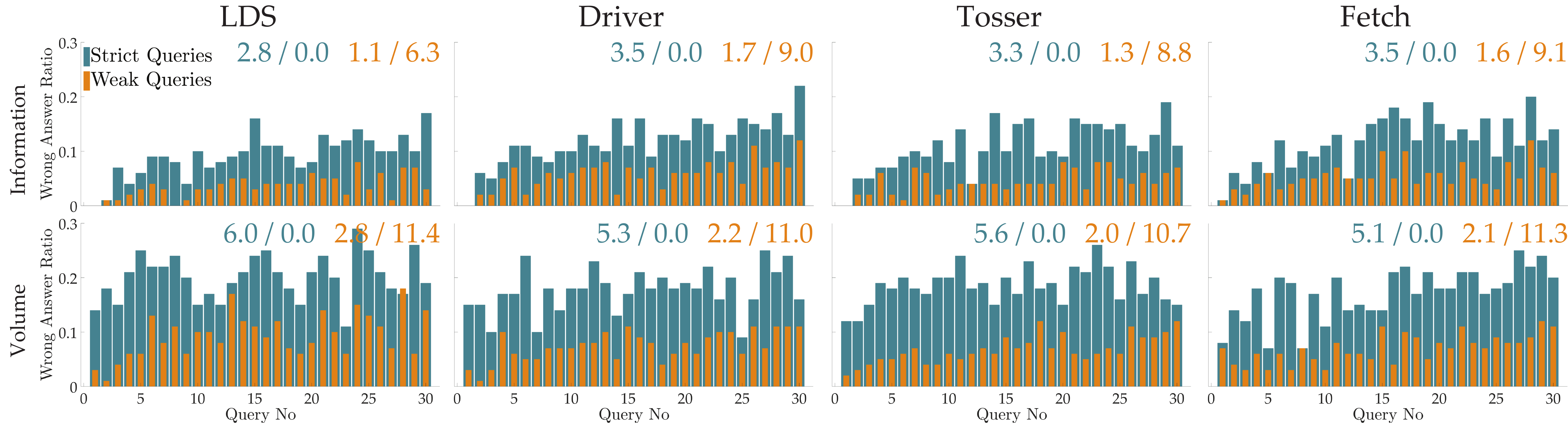}
	\vspace{-15px}
	\caption{Wrong answer ratios on different queries are shown. The numbers at top show the average number of wrong responses and ``About Equal" choices, respectively, for both strict and weak queries.}
	\vspace{-15px}
	\label{fig:simulation_results_wrong_responses}
\end{figure}

The numbers given within Fig.~\ref{fig:simulation_results_wrong_responses} count the wrong answers and ``About Equal" choices made by the simulated users. The information gain formulation significantly improves over volume removal and is slightly better than uniformly random querying. Moreover, weak preference queries consistently decrease the number of wrong answers, which can be one reason why it performs better than strict queries.\footnote{Another possible explanation is the information acquired by the ``About Equal" responses. We analyze this in the Appendix by comparing the results with what would happen if this information was discarded.} Fig.~\ref{fig:simulation_results_wrong_responses} also shows when the wrong responses are given. While wrong answer ratios are higher with volume removal formulation, it can be seen that information gain reduces wrong answers especially in early queries, which leads to faster learning. These results support \textbf{H2}.

\begin{figure}[b]
	\centering
	\vspace{-18px}
	\includegraphics[width=\textwidth]{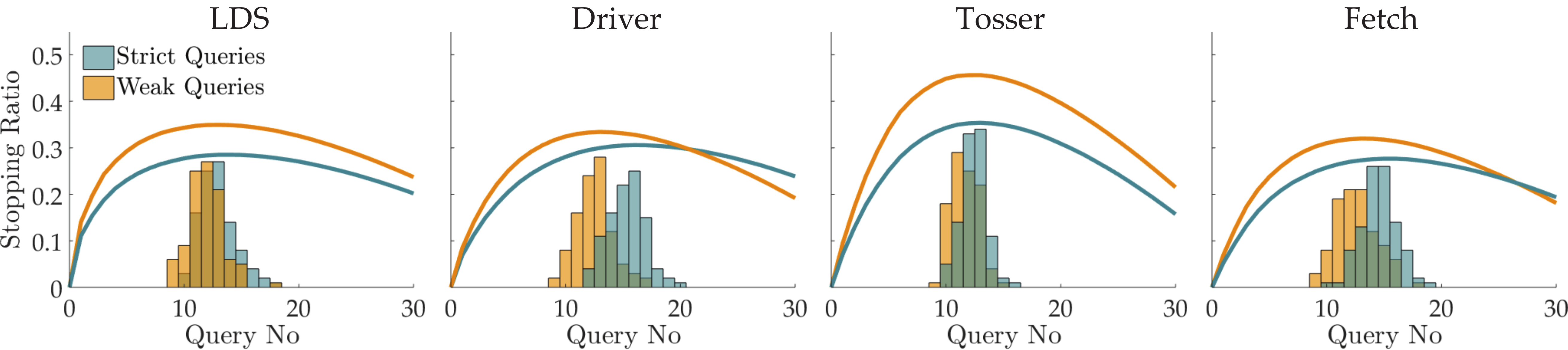}
	\vspace{-15px}
	\caption{Simulation results for optimal stopping. Line plots show cumulative active learning rewards, averaged over $100$ test runs and scaled for visualization. Histograms show when optimal stopping condition is satisfied.}
	\vspace{-5px}
	\label{fig:optimal_stopping_query_dependent}
\end{figure}

\noindent\textbf{Hypothesis 3.} \emph{Optimal stopping enables cost-efficient reward learning under various costs}.
\vspace{-4px}

Lastly, to test \textbf{H3}, we adopted a cost to improve interpretability of queries, which may have the associated benefit of making learning more efficient \cite{bajcsy2018learning}.
We defined a cost function:
\vspace{-3px}
\begin{align}
c(Q) = \epsilon-\abs{\Psi_{i^*}} + \max_{j\in\{1,\dots,d\}\setminus\{i^*\}}\abs{\Psi_j},\: i^*=\argmax_i{\abs{\Psi_i}},
\label{eq:interpretability_cost}
\end{align}
where $Q=\{\xi_1,\xi_2\}$ and $\Psi=\Phi(\xi_1)-\Phi(\xi_2)$.
This cost favors queries in which the difference in one feature is larger than that between all other features. Such a query may prove more interpretable.
We first simulate $100$ random users and tune $\epsilon$ accordingly: For each simulated user, we record the $\epsilon$ value that makes the objective zero in the $i^{\textrm{th}}$ query (for smallest $i$) such that $m_i, m_{i-1}, m_{i-2} \in [x,x+0.02]$ for some $x$. We then use the average of these $\epsilon$ values for our tests with $100$ different random users. Fig.~\ref{fig:optimal_stopping_query_dependent} shows the results.\footnote{We found similar results with query-independent costs minimizing the number of queries. See Appendix.} Optimal stopping rule enables terminating the process with near-optimal cumulative active learning rewards in all environments, which supports \textbf{H3}.

\subsection{User Studies}
\label{sec:user_study}
We deployed our algorithm on a Fetch robot, as well as Driver and Tosser simulations.
We recruited $15$ participants for the simulations and $12$ for the Fetch.
We used strict preference queries; other parameters were the same as in simulations. 
We slightly simplified Fetch feature space and trajectories; see Appendix for details. A video demonstration is available at \url{http://youtu.be/JIs43cO\_g18}.

\textbf{Methodology.} We began by asking participants to rank a set of features (described in plain language) to encourage each user to be consistent in her preferences.
Subsequently, we queried each participant with a  sequence of 30 questions generated actively; 15 from volume removal and 15 via information gain.
We prevent bias by randomizing the sequence of questions for each user and experiment: the user does not know which algorithm generates a question. We test two hypotheses.

\textbf{Hypothesis 4.} \emph{Users find information gain queries easier than those of volume removal.}

Participants responded to a 7-point Likert scale survey after each question: \emph{``It was easy to choose between the trajectories that the robot showed me."} (7-Agree, 1-Disagree). They were also asked the Yes/No question: \emph{``Can you tell the difference between the options presented?"}

\begin{figure}[t]
	\centering
	%\vspace{-10px}
	\includegraphics[width=0.85\textwidth]{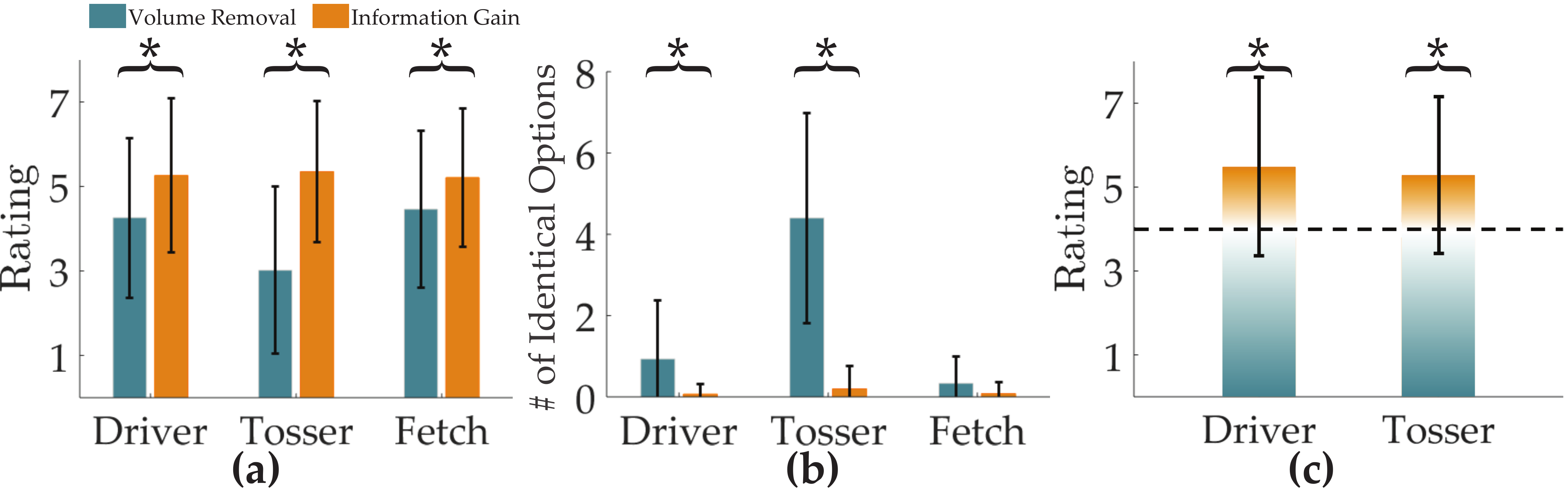}
	\vspace{-8px}
	\caption{User study results. Error bars show std. Asterisks show statistical significance. \textbf{(a)} Easiness survey results averaged over all queries and users. \textbf{(b)} The number of identical options in the experiments averaged over all users. \textbf{(c)} Final preferences averaged over the users. $7$ means the user strongly prefers the optimized trajectory w.r.t. the learned reward by the information gain formulation, and $1$ is the volume removal.}
	\vspace{-12px}
	\label{fig:user_studies}
\end{figure}

Figure~\ref{fig:user_studies}~(a) shows the results of the easiness surveys.
In all environments, users found information gain queries easier; the results are 
statistically significant (two-sample $t$-test, $p<0.005$).
Fig.~\ref{fig:user_studies}~(b) shows the average number of times the users stated they cannot distinguish the options presented.
The volume removal formulation yields several queries that are indistinguishable to the users while the information gain avoids this issue.
The difference is significant for Driver (paired-sample $t$-test, $p<0.05$) and Tosser ($p<0.005$).
Taken together, these results support \textbf{H4}.

\textbf{Hypothesis 5.} \emph{A user's preference aligns best with reward parameters learned via information gain.} 
%\noindent\textbf{H4.} Users find the information gain queries easier than the volume removal queries.\\
%\noindent\textbf{H5.} Users' preferences better align with the reward functions learned with the information gain formulation than with the volume removal.

In concluding the Tosser and Driver experiments, we showed participants two trajectories: one optimized using reward parameters from information gain (trajectory A) and one optimized using reward parameters from volume removal (trajectory B).\footnote{We excluded Fetch for this question to avoid prohibitive trajectory optimization (due to large action space).}
Participants responded to a 7-point Likert scale survey: \emph{``Trajectory A better aligns with my preferences than trajectory B"} (7-Agree, 1-Disagree).

Fig.~\ref{fig:user_studies}(c) shows survey results.
Users significantly preferred the information gain trajectory over that of volume removal in both environments (one-sample $t$-test, $p<0.05$), supporting \textbf{H5}.

\section{Conclusion}
\label{sec:discussion}
\textbf{Summary.} We demonstrate robots can generate both informative and easy queries for reward learning by greedily maximizing their information gain.
This objective trades-off between the uncertainties of the robot and the human about the human's response.
We also use optimal stopping to decide when questions become prohibitively expensive.
Both in simulation and in a user study we demonstrate the benefits of asking the right number of easy questions.
We found our formulation to be more data-efficient and generate easier (as judged by humans) queries than state-of-the-art methods.

\textbf{Limitations and Future Work.} To identify which questions the human can confidently answer, we rely on a cognitive model of the human. Our approach is limited by the accuracy of this model: in future work, we are interested in learning and personalizing this model for the current user. Moreover, it may be sometimes desirable to make some questions explicitly easier. One can use the interpretability cost (Eq.~\eqref{eq:interpretability_cost}), or add a weight term to Eq.~\eqref{eq:easiness_equation} for which further research is warranted.

\vspace{-5px}
\acknowledgments{\vspace{-5px}This work is supported by FLI grant RFP2-000 and NSF Award \#1849952. Toyota Research Institute (TRI) provided funds to assist the authors with their research but this article solely reflects the opinions and conclusions of its authors and not TRI or any other Toyota entity.}

%===============================================================================

\clearpage
\setlength{\bibsep}{4.70pt plus 0.3ex}
\small\bibliography{refs}  % .bib
\clearpage
\normalsize

\section{Appendix}
\subsection{Derivation of Information Gain Solution}
\label{sec:info_gain_derivation}
\begin{align*}
Q^*_n &= \argmax_{Q_n=\{\Delta_1,\dots,\Delta_K\}} I(q_n ; \omega \mid  Q_n)\\
I(q_n& ; \omega \mid  Q_n)\\
&= H(\omega \mid  Q_n) - \mathbb{E}_{q_n\mid Q_n}\left[H(\omega \mid  q_n,Q_n)\right]\\
&= -\mathbb{E}_{\omega\mid Q_n}\left[\log_2\left(P(\omega\mid Q_n)\right)\right] + \mathbb{E}_{\omega,q_n\mid Q_n}\left[\log_2\left(P(\omega \mid  q_n,Q_n)\right)\right]\\
&=  \mathbb{E}_{\omega,q_n\mid Q_n}\left[\log_2\left(P(\omega \mid  q_n,Q_n)\right)-\log_2\left(P(\omega\mid Q_n)\right)\right]\\
&= \mathbb{E}_{\omega, q_n\mid Q_n}\left[\log_2\left(P(q_n \mid  Q_n,\omega)\right)-\log_2\left(P(q_n\mid Q_n)\right)\right]\\
&= \mathbb{E}_{\omega, q_n\mid Q_n}\left[\log_2\left(P(q_n \mid  Q_n,\omega)\right)-\log_2\left(\int P(q_n\mid Q_n,\omega')P(\omega' \mid  Q_n)d\omega'\right)\right]\\
&\asymeq \mathbb{E}_{\omega, q_n\mid Q_n}\left[\log_2\left(P(q_n \mid  Q_n,\omega)\right)-\log_2\left(\frac1M \sum_{\omega'\in\Omega} P(q_n\mid Q_n,\omega')\right)\right]\\
&= \mathbb{E}_{\omega, q_n\mid Q_n}\left[\log_2\left(\frac{M\cdot P(q_n \mid  Q_n,\omega)}{ \sum_{\omega'\in\Omega} P(q_n\mid Q_n,\omega')}\right)\right]\\
&= \mathbb{E}_{\omega\mid Q_n}\left[\mathbb{E}_{q_n\mid Q_n,\omega}\left[\log_2\left(\frac{M\cdot P(q_n \mid  Q_n,\omega)}{ \sum_{\omega'\in\Omega} P(q_n\mid Q_n,\omega')}\right)\right]\right]\\
&= \mathbb{E}_{\omega\mid Q_n}\left[\sum_{q_n\in Q_n} P(q_n\mid Q_n,\omega)\log_2\left(\frac{M\cdot P(q_n \mid  Q_n,\omega)}{ \sum_{\omega'\in\Omega} P(q_n\mid Q_n,\omega')}\right)\right]\\
&\asymeq \frac1M \sum_{q_n\in Q_n}\sum_{\bar{\omega}\in\Omega} P(q_n\mid Q_n,\bar{\omega})\log_2\left(\frac{M\cdot P(q_n \mid  Q_n,\bar{\omega})}{ \sum_{\omega'\in\Omega} P(q_n\mid Q_n,\omega')}\right)
\end{align*}
where the integral is taken over all possible values of $\omega$.

\subsection{Proof of Theorem 3}
\setcounter{theorem}{2}
\begin{theorem}
	Terminating the algorithm is optimal if and only if global solution to \eqref{eq:stopping_opt} is negative.
	\vspace{-10px}
	\begin{proof}
	 	We need to show if the global optimum is negative, then any longer-horizon optimization will also give negative reward in expectation. Let $Q^*_n$ denote the global optimizer. For any $k\geq0$,
	 	\begin{align*}
	 	I(&q_n,\dots,q_{n+k} ; \omega \mid  Q_n,\dots,Q_{n+k}) - \sum_{i=0}^{k}c(Q_{n+i}) \\
	 	&= I(q_n ; \omega \mid  Q_n)\! +\! \mydots\! +\! I(q_{n+k} ; \omega \mid  q_n, \mydots, q_{n+k-1}, Q_n,\mydots,Q_{n+k})\! -\! \sum_{i=0}^{k}c(Q_{n+i})\\
	 	&\leq I(q_n ; \omega \mid  Q_n)\! +\! \mydots\! +\! I(q_{n+k} ; \omega \mid  Q_{n+k})\! -\! \sum_{i=0}^{k}c(Q_{n+i})\\
	 	&\leq (k+1)\left[I(q_n ; \omega \mid  Q^*_n) - c(Q^*_n)\right] < 0
	 	\end{align*}
	 	where the first inequality is due to the submodularity of the mutual information, and the second inequality is because $Q^*_n$ is the global maximizer of the greedy objective. The other direction of the proof is very clear: If the global optimizer is nonnegative, then querying $Q^*_n$ will not decrease the cumulative active learning reward in expectation, so stopping is not optimal.
	\end{proof}
\end{theorem}

\subsection{Extension to User-Specific and Unknown $\delta$}
We now derive the information gain solution when the parameter $\delta$ of the human model we introduced in Sec.~\ref{sec:human_model} is unknown. One can also introduce a temperature parameter $\beta$ to the softmax model such that $R(\xi_k)$ values will be replaced with $\beta R(\xi_k)$ in Eqs.~\eqref{eq:noisy_rational} and \eqref{eq:weak_noisy_rational}. This temperature parameter is useful for setting how noisy the user choices are, and learning it can ease the feature design.

Therefore, for generality, we denote all human model parameters that will be learned as a vector $\nu$. Since our true goal is to learn $\omega$, the optimization now becomes:
\begin{align*}
Q_n^* &= \argmax_{Q_n=\{\Lambda_1,\dots,\Lambda_K\}} \mathbb{E}_{\nu\mid Q_n}\left[I(q_n;\omega \mid  Q_n)\right]
\end{align*}
We now work on this objective as we did in Sec.~\ref{sec:info_gain_derivation}:
\begin{align*}
\mathbb{E}&_{\nu\mid Q_n}\left[I(q_n;\omega \mid  Q_n)\right]\\
&= \mathbb{E}_{\nu\mid Q_n}\left[H(\omega \mid  \nu, Q_n) - \mathbb{E}_{q_n\mid \nu, Q_n}\left[H(\omega \mid  q_n, \nu, Q_n)\right]\right]\\
&= \mathbb{E}_{\nu\mid Q_n}\left[H(\omega \mid  \nu, Q_n)\right] - \mathbb{E}_{\nu,q_n\mid Q_n}\left[H(\omega \mid  q_n, \nu, Q_n)\right]\\
&= -\mathbb{E}_{\nu,\omega\mid Q_n}\left[\log_2\left(P(\omega \mid  \nu, Q_n)\right)\right] + \mathbb{E}_{\nu,q_n,\omega\mid Q_n}\left[\log_2\left(P(\omega \mid  q_n, \nu, Q_n)\right)\right]\\
&= \mathbb{E}_{\nu,q_n,\omega\mid Q_n}\left[\log_2\left(P(\omega \mid  q_n, \nu, Q_n)\right) - \log_2\left(P(\omega \mid  \nu, Q_n)\right)\right]\\
&= \mathbb{E}_{\nu,q_n,\omega\mid Q_n}\left[\log_2\left(P(q_n \mid  \omega, \nu, Q_n)\right) - \log_2\left(P(q_n \mid  \nu, Q_n)\right)\right]\\
&= \mathbb{E}_{\nu,q_n,\omega\mid Q_n}\!\left[\log_2\left(P(q_n \mid  \omega, \nu, Q_n)\right)\!-\! \log_2\left(P(\nu,q_n \mid  Q_n)\right)\!+\!\log_2\left(P(\nu \mid  Q_n)\right)\right]
\end{align*}
Noting that $P(\nu \mid  Q_n) = P(\nu)$, we drop the last term because it does not involve the optimization variable $Q_n$. Then, the new objective is:
\begin{align*}
& \mathbb{E}_{\nu,q_n,\omega\mid Q_n}\!\left[\log_2\left(P(q_n \mid  \omega, \nu, Q_n)\right)\!-\! \log_2\left(P(\nu,q_n \mid  Q_n)\right)\right]\\
&\asymeq \frac1M \sum_{(\bar{\omega},\bar{\nu}) \in \Omega^+}\sum_{q_n\in Q_n}P(q_n\mid \bar{\omega},\bar{\nu},Q_n)\left[\log_2\left(P(q_n \mid  \bar{\omega}, \bar{\nu}, Q_n)\right)\!-\! \log_2\left(P(\bar{\nu},q_n \mid  Q_n)\right)\right]
\end{align*}
where $\Omega^+$ is a set that contains $M$ samples from $P(\omega,\nu)$. Since $P(\bar{\nu},q_n \mid  Q_n) = \int P(q_n \mid  \bar{\nu}, \omega', Q_n)P(\bar{\nu},\omega' \mid  Q_n)d\omega'$ where the integration is over all possible values of $\omega$, we can write the second logarithm term as:
\begin{align*}
\log_2\left(\frac1M \sum_{\omega'\in \Omega(\bar{\nu})}P(q_n \mid  \bar{\nu},\omega', Q_n)\right)
\end{align*}
with asymptotic equality, where $\Omega(\bar{\nu})$ is the set that contains $M$ samples from $P(\omega,\bar{\nu})$ with fixed $\bar{\nu}$. Note that while we can actually compute this objective, it is computationally much heavier than the case without $\nu$, because we need to take $M$ samples of $\omega$ for each $\bar{\nu}$ sample.

One property of this objective that will ease the computation is the fact that it is parallelizable. An alternative approach is to actively learn $(\omega,\nu)$ instead of just $\omega$. This will of course cause some performance loss, because we are only interested in $\omega$. However, if we learn them together, the derivation follows Sec.~\ref{sec:info_gain_derivation} by simply replacing $\omega$ with $(\omega,\nu)$, and the final optimization becomes:
\begin{align*}
\argmax_{Q_n=\{\Lambda_1,\dots,\Lambda_K\}} \frac1M \sum_{q_n\in Q_n}\sum_{(\bar{\omega},\bar{\nu})\in\Omega^+} P(q_n\mid Q_n,\bar{\omega},\bar{\nu})\log_2\left(\frac{M\cdot P(q_n \mid  Q_n,\bar{\omega},\bar{\nu})}{ \sum_{(\omega',\nu')\in\Omega^+} P(q_n\mid Q_n,\omega',\nu')}\right)
\end{align*}

\begin{figure}[t]
	\centering
	\includegraphics[width=\textwidth]{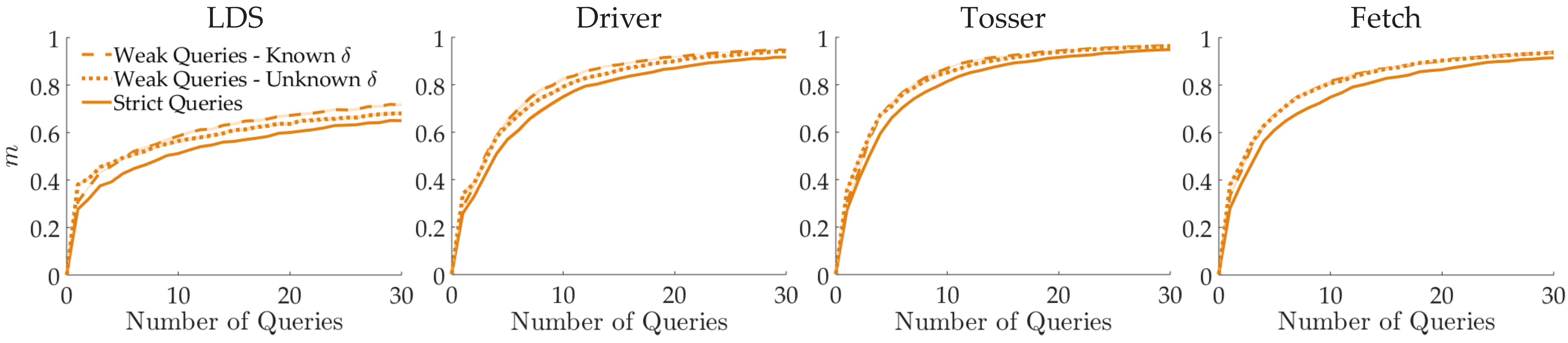}
	\vspace{-15px}
	\caption{The simulation results with information gain formulation for unknown $\delta$. Plots are mean$\pm$s.e.}
	\vspace{-10px}
	\label{fig:unknown_delta}
\end{figure}

Using this approximate, but computationally faster  optimization, we performed additional analysis where we compare the performances of strict preference queries, weak preference queries with known $\delta$ and weak preference queries without assuming any $\delta$ (all with the information gain formulation). As in the previous simulations, we simulated $100$ users with different random reward functions. Each user is simulated to have a true delta, uniformly randomly taken from $[0,2]$. During the sampling of $\Omega^+$, we did not assume any prior knowledge about $\delta$, except the natural condition that $\delta\geq 0$. The comparison results are in Fig.~\ref{fig:unknown_delta}. While knowing $\delta$ increases the performance as it is expected, weak preference queries are still better than strict queries even when $\delta$ is unknown. This supports the advantage of employing weak queries.

\subsection{Feature Transformations and Action Spaces of the Environments}
\label{sec:fetch_modifications}
We give the details about the simulation environments in Table~\ref{tab:environment_details}. We slightly modified the Fetch task for the user studies. We removed the average speed feature and set the robot such that it directly moves to its final position without following each of $19$ control inputs.

\begin{table}[b]
	\caption{Environment Details}
	\label{tab:environment_details}
	\centering
	\begin{tabular}{cccc}
		\hline
		& \textbf{Driver} & \textbf{Tosser} & \textbf{Fetch (Simulations)} \\ \hline
		\textbf{Features} & \begin{minipage}{105pt}
			\begin{itemize}[nosep]
				\item The average of $e^{-c_1d_1^2}$ over the trajectory, where $d_1$ is the shortest distance between the ego car and a lane center, and $c_1\!=\!30$.
				\item The average of $(v_1\!-\!1)^2$ over the trajectory, where $v_1$ is the speed of the ego car.
				\item The average of $\cos(\theta_1)$ over the trajectory, where $\theta_1$ is the angle between the directions of ego car and the road.
				\item The average of $e^{-c_2d_2^2-c_3d_3^2}$ over the trajectory, where $d_2$ and $d_3$ are the horizontal and vertical distances between the ego car and the other car, respectively; and $c_2\!=\!7$, $c_3\!=\!3$.
			\end{itemize}
			\vspace{5px}
		\end{minipage} &
		\begin{minipage}{105pt}
			\begin{itemize}[nosep]
				\item The maximum distance the object moved forward from the tosser.
				\item The maximum altitude of the object.
				\item Number of flips (real number) the object does.
				\item $e^{-c_4d_4}$ where $d_4$ is the final horizontal distance between the object and the center of the closest basket, and $c_4=3$.
			\end{itemize}
			\vspace{5px}
		\end{minipage} & 
		\begin{minipage}{105pt}
			\begin{itemize}[nosep]
				\item The average of $e^{-c_5d_5}$ over the trajectory where $d_5$ is the distance between the end effector and the goal object, and $c_5=1$.
				\item The average of $e^{-c_6d_6}$ over the trajectory where $d_6$ is the vertical distance between the end effector and the table, and $c_6=1$.
				\item The average of $e^{-c_7d_7}$ over the trajectory where $d_7$ is the distance between the end effector and the obstacle, and $c_7=1$.
				\item The average of the end effector speed over the trajectory.
			\end{itemize}
			\vspace{5px}
		\end{minipage}\\
		\hline
		$\stackrel{\hbox{\textbf{Action}}}{\hbox{\textbf{Space}}}$ & 
		\begin{minipage}{105pt}
			\vspace{5px}
			The ego car is given steering and acceleration values $5$ times, each of which is applied for $10$ time steps.
			\vspace{5px}
		\end{minipage} &
		\begin{minipage}{105pt}
			\vspace{5px}
			The two joints of the tosser are given control inputs twice. Both the first and the second set of inputs are applied for $25$ time steps each, after the object is in free fall for $150$ time steps. In the next $800$ time steps, the user of the system observes the object's trajectory.
			\vspace{5px}
		\end{minipage} &
		\begin{minipage}{105pt}
			\vspace{5px}
			The orientation of the gripper is fixed such that the end effector always points down. For the position of the gripper, we give control inputs $19$ times, each of which is applied for $8$ time steps.
			\vspace{5px}
		\end{minipage} \\
		\hline
	\end{tabular}
\end{table}

\subsection{Comparison of Models without Query Space Discretization}
\begin{figure}[t]
	\centering
	\includegraphics[width=0.75\textwidth]{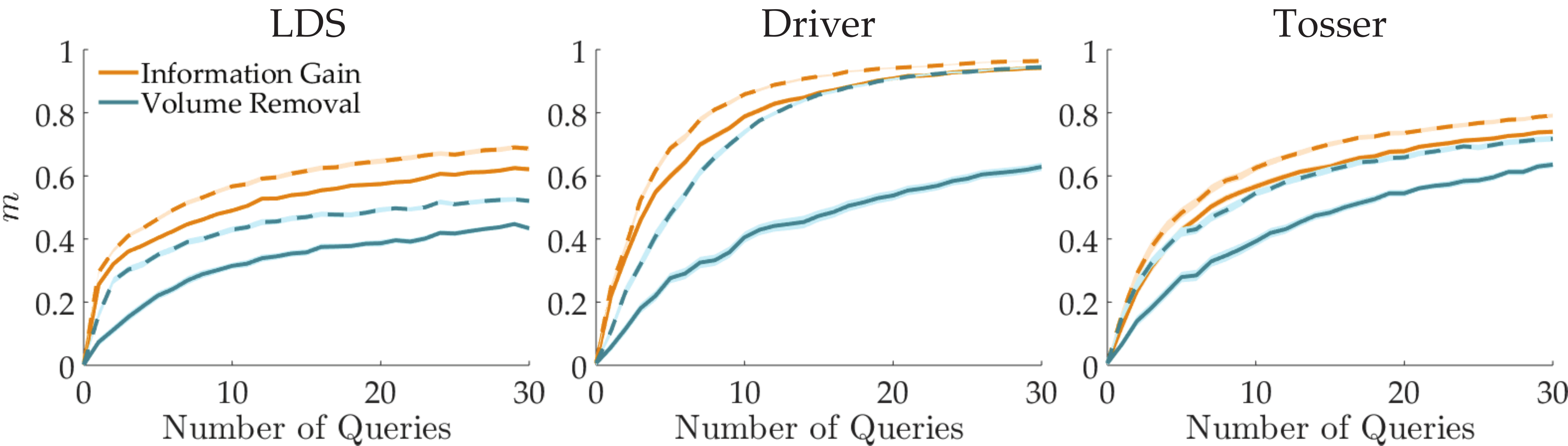}
	\vspace{-5px}
	\caption{Alignment values are plotted (mean$\pm$s.e.). Dashed lines show the weak preference query variants.}
	\vspace{-15px}
	\label{fig:continuous_optimization}
\end{figure}

We repeated the experiment that supports \textbf{H1}, and whose results are shown in Fig.~\ref{fig:simulation_results_m}, without query space discretization. By optimizing over the continuous action space of the environments, we tested information gain and volume removal formulations with both strict and weak preference queries in LDS, Driver and Tosser tasks. We excluded Fetch again in order to avoid prohibitive trajectory optimization due to large action space. Fig.~\ref{fig:continuous_optimization} shows the result. As it is expected, information gain formulation outperforms the volume removal with both preference query types. And, weak preference queries lead to faster learning compared to strict preference queries.

\subsection{Effect of Information from ``About Equal" Responses}
We have seen that weak preference queries consistently decrease wrong answers and improve the performance. However, this improvement is not necessarily merely due to the decrease in wrong answers. It can also be credited to the information we acquire thanks to ``About Equal" responses.

\begin{figure}[h]
	\centering
	\vspace{-10px}
	\includegraphics[width=\textwidth]{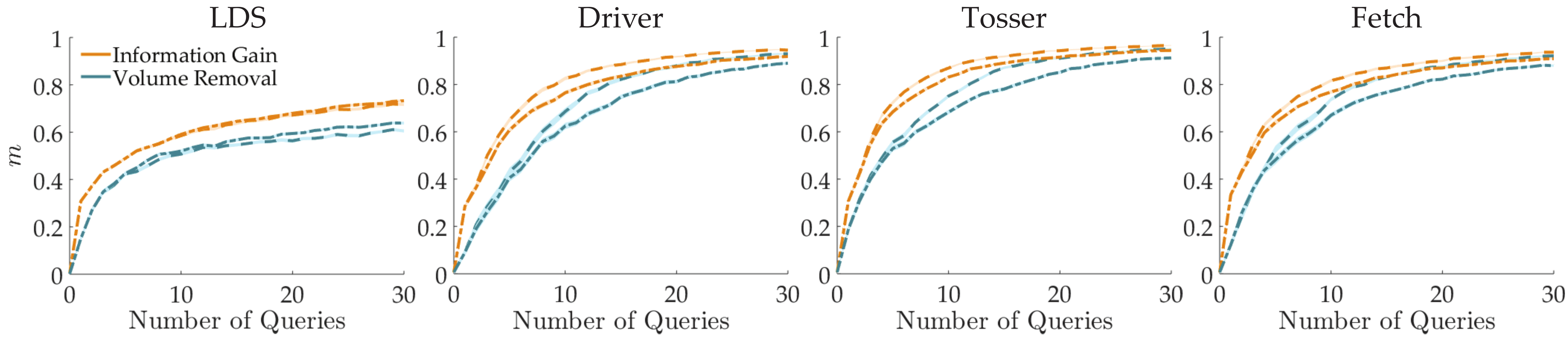}
	\vspace{-15px}
	\caption{The results (mean$\pm$standard error) of the simulations with weak preference queries where we use the information from ``About Equal" responses (dashed lines) and where we don't use (dash-dot lines).}
	\vspace{-10px}
	\label{fig:value_of_idk}
\end{figure}

To investigate the effect of this information, we perform two additional experiments with $100$ different simulated human reward functions with weak preference queries: First, we use the information by the ``About Equal" responses; and second, we ignore such responses and remove the query from the query set to prevent repetition. We again take $M=100$ and use Metropolis-Hastings algorithm for sampling. Fig.~\ref{fig:value_of_idk} shows the results. It can be seen that for both volume removal and information gain formulations, the information from ``About Equal" option improves the learning performance in Driver, Tosser and Fetch tasks, whereas its effect is very small in LDS.

\subsection{Optimal Stopping under Query-Independent Costs}
\begin{figure}[t]
	\centering
	\includegraphics[width=\textwidth]{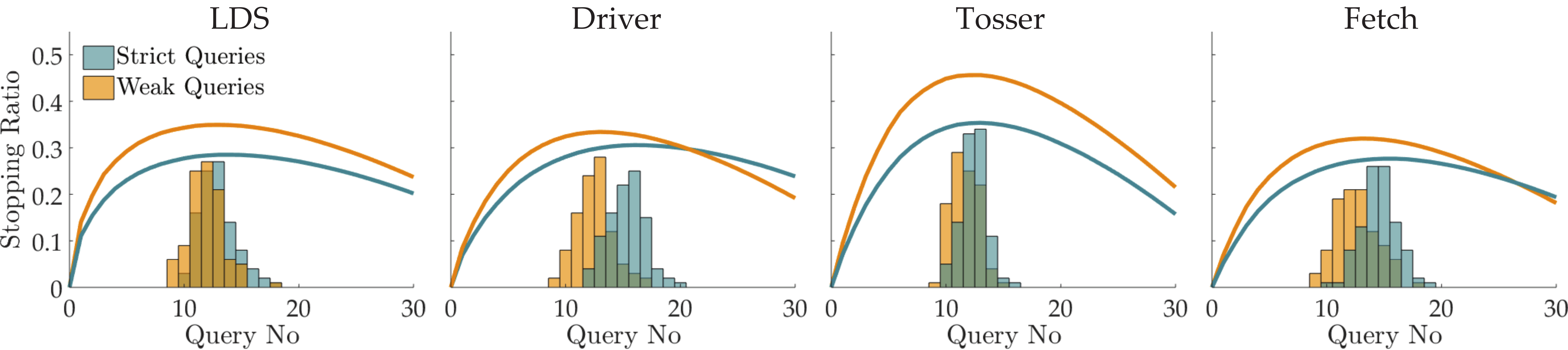}
	\vspace{-15px}
	\caption{Simulation results for optimal stopping under query-independent costs. Line plots show cumulative active learning rewards, averaged over $100$ test runs and scaled for better appearance. Histograms show when optimal stopping condition is satisfied.}
	\vspace{-20px}
	\label{fig:optimal_stopping_query_independent}
\end{figure}

To investigate optimal stopping performance under query-independent costs, we defined the cost function as $c(Q) = \epsilon$, which just balances the trade-off between the number of questions and learning performance. Similar to the query-dependent costs case we described in \ref{sec:simulations}, we first simulate $100$ random users and tune $\epsilon$ accordingly: For each simulated user, we record the $\epsilon$ value that makes the objective zero in the $i^{\textrm{th}}$ query such that $m_i, m_{i-1}, m_{i-2} \in [x,x+0.02]$ for some $x$. We then use the average of these $\epsilon$ values for our tests with $100$ different random users. Fig.~\ref{fig:optimal_stopping_query_independent} shows the results. Optimal stopping rule enables terminating the process with near-optimal cumulative active learning rewards in all environments, which again supports \textbf{H3}.

\end{document}